\documentclass{article}

\usepackage{microtype}
\usepackage{graphicx}
\usepackage{booktabs}
\usepackage{hyperref}
\usepackage[accepted]{icml2026}
\usepackage{amsmath}
\usepackage{amssymb}
\usepackage{enumitem}
\usepackage{xcolor}

\newcommand{\method}[1]{\textsc{#1}}
\newcommand{\metric}[1]{\texttt{#1}}

\newcommand{\err}[1]{{\scriptsize$\pm$#1}}

\begin{document}

\twocolumn[
\icmltitle{Theory-Grounded Evaluation Exposes the Authorship Gap\texorpdfstring{\\}{ }in LLM Personalization}

\begin{icmlauthorlist}
\icmlauthor{Yash Ganpat Sawant}{ind}
\end{icmlauthorlist}
\icmlaffiliation{ind}{Independent AI Researcher}
\icmlcorrespondingauthor{Yash Ganpat Sawant}{sawantyash13@gmail.com}

\vskip 0.3in
]

\printAffiliationsAndNotice{}

\icmltitlerunning{Theory-Grounded Evaluation Exposes the Authorship Gap in LLM Personalization}

\begin{abstract}

Stylistic personalization---making LLMs write in a specific individual's style, rather than merely adapting to task preferences---lacks evaluation grounded in authorship science.
We show that grounding evaluation in authorship verification theory transforms what benchmarks can measure.
Drawing on three measurement traditions---LUAR (a trained authorship verification model), an LLM-as-judge with decoupled trait matching, and classical function-word stylometrics---we evaluate four inference-time personalization methods across 50 authors and 1{,}000 generations.
The theory-grounded metric (LUAR) provides what ad hoc alternatives cannot: calibrated baselines (human ceiling 0.756, cross-author floor 0.626) that give scores absolute meaning.
All methods score \emph{below} this floor (0.484--0.508), exposing an authorship gap invisible to uncalibrated metrics.
The three metrics produce near-zero pairwise correlations ($|r| < 0.07$), confirming that without theoretical grounding, metric choice determines conclusions---an LLM judge declares a clear winner while LUAR finds no meaningful differentiation.
These findings demonstrate the theory--benchmark cycle in action: authorship theory exposes evaluation failures that ad hoc benchmarks miss.

\end{abstract}

\section{Introduction}
\label{sec:intro}

Stylistic personalization---making an LLM write in a specific individual's style, rather than merely adapting to task preferences---has no established evaluation methodology grounded in authorship science.
LaMP~\citep{salemi2024lamp} evaluates personalization through task accuracy.
PersonalLLM~\citep{personalllm2025} measures preference alignment.
PersonaLens~\citep{zhao2025personalens} uses LLM-as-judge for conversational personalization.
None evaluate whether generated text \emph{sounds like} the target author---whether the model's underlying authorship fingerprint actually shifts toward the target.

This is a gap, not a critique: authorial style fidelity is simply a different construct than task accuracy or preference alignment, and no standard metric for it exists.
We propose to fill this gap by grounding evaluation in authorship verification theory---a decades-old discipline with validated methods and calibrated baselines~\citep{stamatatos2009survey, riverasoto2021luar}.
We contrast this theory-grounded approach with two ad hoc alternatives: an LLM-as-judge (a decoupled binary trait protocol) and classical stylometrics (function word distributions~\citep{argamon2003style}).

We test these three metrics on four inference-time stylistic personalization methods across 50 authors and 1{,}000 generations, and find:

\begin{enumerate}[leftmargin=*, itemsep=2pt]
    \item \textbf{Theory provides calibrated baselines that ad hoc metrics lack.} LUAR authorship verification yields a human-author ceiling (0.756) and cross-author floor (0.626), giving scores absolute meaning. All methods score \emph{below} the human floor (0.484--0.508), exposing an authorship gap invisible without calibration.
    \item \textbf{Without theoretical grounding, metric choice determines conclusions.} The LLM judge declares profile extraction a clear winner ($d{=}0.58$); LUAR finds no meaningful differentiation. The judge's apparent signal traces to circularity between trait extraction and profile extraction (Section~\ref{sec:circularity}).
    \item \textbf{Metric disagreement signals construct validity failure.} The three metrics produce near-zero correlations ($|r| < 0.07$)---they measure different constructs, and only the theory-grounded metric passes validation tests.
\end{enumerate}

These findings demonstrate the theory--benchmark cycle: authorship theory generates testable predictions, the benchmark confirms them with calibrated measurements, and the result exposes evaluation failures that ad hoc approaches miss.

\section{Related Work}
\label{sec:related}

\paragraph{Personalization benchmarks.}
LaMP~\citep{salemi2024lamp} evaluates personalization via task accuracy; LongLaMP~\citep{kumar2024longlamp} extends this to long-form generation with content-summary prompts (which we adopt).
PersonalLLM~\citep{personalllm2025} uses synthetic preference profiles.
PersonaLens~\citep{zhao2025personalens} evaluates conversational personalization with LLM-as-judge.
Critically, no existing benchmark evaluates whether generated text is stylistically faithful to the target author---the gap we address.

\paragraph{Benchmark quality and construct validity.}
BetterBench~\citep{betterbench2024} proposes 46 criteria for assessing benchmark quality, finding widespread disparities.
\citet{raji2021benchmark} argue that benchmarks become proxies for progress without validating whether they measure the intended construct---an instance of the broader construct validity problem~\citep{cronbach1955construct}.
Our work applies this lens specifically to personalization: we propose three metrics from different traditions, show they diverge, and identify which one---LUAR authorship verification---passes validation tests the others fail.

\paragraph{Authorship verification.}
Computational authorship analysis spans from function-word frequencies~\citep{argamon2003style} to neural methods.
LUAR~\citep{riverasoto2021luar} learns universal authorship representations via contrastive learning.
\citet{wang2025catchme} use authorship analysis to show LLMs struggle to imitate everyday authors---a finding our calibrated baselines quantify precisely.

\section{Evaluation Framework}
\label{sec:framework}

\subsection{Data and Methods}

We use the Blog Authorship Corpus~\citep{schler2006effects}: 681K posts from 19,320 bloggers.
We select 50 authors with $\geq$200 training posts, $\geq$50 test posts, and mean length $\geq$100 words, yielding 104K training and 26K test posts.
Writing prompts are LLM-extracted content summaries (neutral descriptions of \emph{what} a post discusses, not \emph{how})---we show in Section~\ref{sec:circularity} that na\"ive first-sentence extraction inflates baselines by 28 percentage points.

We evaluate four inference-time methods spanning implicit to explicit style transfer: \method{Non-Personalized} (control; content summary only), \method{Few-Shot} (5 author samples, no explicit instruction), \method{Profile Extraction} (two-stage: extract abstract style profile, then generate from profile only), and \method{Contrastive} (author samples + contrastive examples from other authors + stylometric features).
All use Qwen~3 32B~\citep{qwen2025qwen3} as generator with 50 authors $\times$ 5 prompts $\times$ 4 methods $= 1{,}000$ generations.

\subsection{Three Independent Metrics}
\label{sec:metrics}

We deliberately select metrics from three different traditions:

\paragraph{LUAR (Primary).}
Learning Universal Authorship Representations~\citep{riverasoto2021luar} is a transformer trained via contrastive learning on millions of Reddit posts to produce author-discriminative embeddings.
We compute cosine similarity between 5-post aggregated LUAR embeddings of generated and real text.
LUAR is uniquely suited to personalization evaluation because it provides \emph{calibrated} baselines: same-author pairs yield known score distributions distinct from cross-author pairs, enabling absolute rather than relative evaluation.

We validate LUAR on our blog corpus (trained on Reddit): single-post AUC$=$0.76, multi-post (5) AUC$=$0.96, vs.\ TF-IDF baseline AUC$=$0.54.
The metric transfers reliably across domains.

\paragraph{LLM-as-Judge (Secondary).}
A decoupled binary protocol using GLM-4 32B~\citep{glm2024chatglm} (different model family from generator):
(i)~extract 5 style traits as yes/no questions per author (cached);
(ii)~score each generation on traits in one call;
(iii)~judge same-author plausibility in a \emph{separate} call.
Decoupling prevents cross-signal contamination between trait scoring and holistic judgment.
Primary metric: \metric{Trait Match Rate} (TMR) $= \text{traits\_present}/5$.

\paragraph{Automated Stylometrics (Tertiary).}
Function word cosine similarity (\metric{FuncCos}) over 60 common function words---established markers of individual writing style~\citep{argamon2003style}.

\section{Results}
\label{sec:results}

\subsection{The Human--LLM Authorship Gap}

\begin{table}[t]
\centering
\caption{Method comparison across 50 authors, 1{,}000 generations. LUAR uses 5-post aggregation ($\uparrow$ better). TMR = trait match rate. SA\% = same-author rate. All methods score \emph{below} the cross-author human floor on LUAR. CIs: hierarchical bootstrap ($B{=}10{,}000$).}
\label{tab:main}
\small
\resizebox{\columnwidth}{!}{%
\begin{tabular}{lcccc}
\toprule
\textbf{Method} & \textbf{LUAR} $\uparrow$ & \textbf{TMR} $\uparrow$ & \textbf{SA\%} $\uparrow$ & \textbf{FuncCos} $\uparrow$ \\
\midrule
\method{Non-Personal.} & 0.484\err{.019} & 0.384\err{.058} & 22\% & 0.741\err{.011} \\
\method{Few-Shot}       & \textbf{0.508}\err{.020} & 0.433\err{.061} & 31\% & 0.749\err{.011} \\
\method{Profile Extr.}  & 0.502\err{.019} & \textbf{0.542}\err{.060} & 29\% & \textbf{0.761}\err{.010} \\
\method{Contrastive}    & 0.494\err{.020} & 0.447\err{.059} & \textbf{36\%} & 0.752\err{.011} \\
\midrule
Real Author (ceil.)     & 0.756 & 0.427 & 30\% & 0.742 \\
Cross-Author (floor)    & 0.626 & 0.390 & 7\% & 0.695 \\
\bottomrule
\end{tabular}%
}
\end{table}

Table~\ref{tab:main} presents our central measurement.
On LUAR, all four methods score between 0.484 and 0.508---a spread of just \textbf{0.024}---and all fall \emph{below} the human cross-author floor of 0.626 (ceiling 0.756).
The LLM's authorship fingerprint dominates: personalized output is more distant from the target human author than random humans are from each other.

\begin{figure}[t]
\centering
\includegraphics[width=\columnwidth]{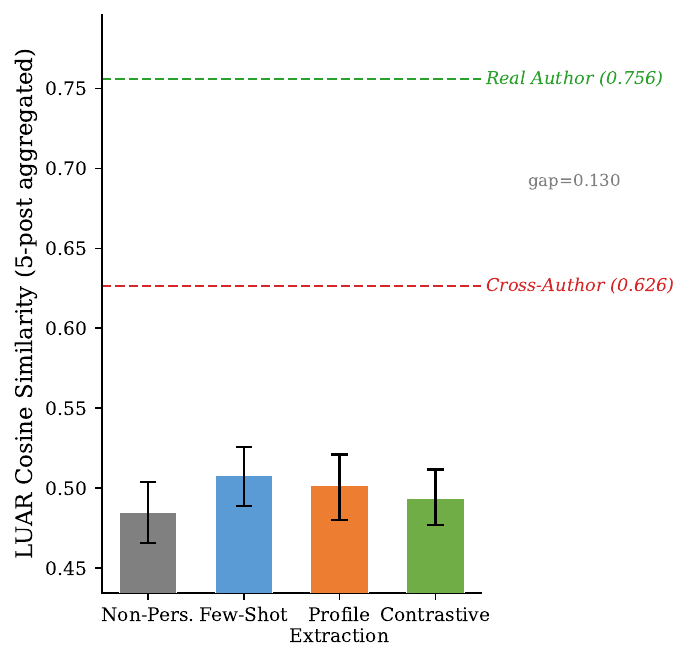}
\caption{LUAR authorship similarity by method with calibration baselines. All methods score below the cross-author human floor (0.626). The authorship gap between generated and human text is a measurable, calibrated quantity.}
\label{fig:luar_main}
\end{figure}

Yet personalization is not vacuous.
Within generated text, LUAR discriminates target authors at AUC$=$0.918 (gen$\leftrightarrow$gen), confirming that methods produce author-differentiated output.
This output simply remains in the LLM's own style space rather than crossing into human authorship territory (gen$\rightarrow$real AUC$=$0.632).
Figure~\ref{fig:luar_main} visualizes the gap.

\subsection{Why LUAR Should Anchor Personalization Evaluation}

\begin{table}[t]
\centering
\caption{Pearson correlation between metrics ($n{=}1{,}000$). All correlations near zero: the metrics capture fundamentally different constructs.}
\label{tab:correlations}
\small
\begin{tabular}{lccc}
\toprule
& \textbf{LUAR} & \textbf{TMR} & \textbf{FuncCos} \\
\midrule
\textbf{LUAR}    & 1.00  & ---   & --- \\
\textbf{TMR}     & 0.013 & 1.00  & --- \\
\textbf{FuncCos} & 0.026 & 0.067 & 1.00 \\
\bottomrule
\end{tabular}
\end{table}

Table~\ref{tab:correlations} reveals that neither TMR nor FuncCos correlates with LUAR---the only metric validated against known authorship baselines ($|r| < 0.07$; bootstrap 95\% CIs: LUAR--TMR $[-0.049, 0.075]$, LUAR--FuncCos $[-0.036, 0.089]$).
This means a benchmark using only TMR would declare \method{Profile Extraction} the clear winner (effect size $d{=}0.58$ over baseline), while LUAR---anchored to calibrated authorship verification---finds no meaningful differentiation across methods.

\begin{figure}[t]
\centering
\includegraphics[width=\columnwidth]{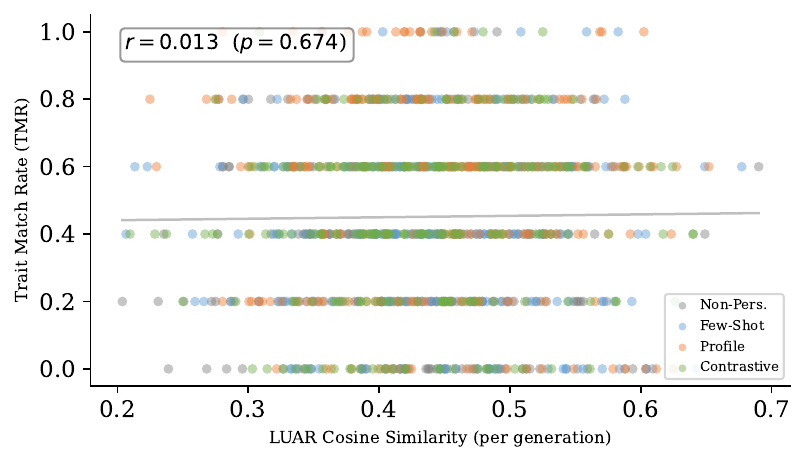}
\caption{LUAR similarity vs.\ TMR for 1{,}000 generations ($r{=}0.013$). Profile extraction's apparent advantage on TMR has no corresponding signal on LUAR.}
\label{fig:cross_metric}
\end{figure}

Why trust LUAR over the alternatives? LUAR's baselines are well-separated (ceiling 0.756, floor 0.626, gap$=$0.130), confirming it discriminates authors reliably (AUC$=$0.96, Section~\ref{sec:metrics}).
FuncCos baselines are nearly collapsed (ceiling 0.742, floor 0.695, gap$=$0.047)---it cannot meaningfully separate same-author from cross-author text, and generated methods (0.741--0.761) straddle the ceiling, suggesting LLMs produce grammatically average function-word distributions regardless of personalization.
TMR has no calibrated baselines at all, and the circularity evidence from Section~\ref{sec:circularity} shows \method{Profile Extraction} \emph{exceeds the real author} on TMR (0.542 vs.\ 0.427)---a method that scores higher than ground truth is measuring instruction-following, not authorship fidelity.
Figure~\ref{fig:cross_metric} visualizes this: TMR scores are uniformly distributed across the LUAR range ($r{=}0.013$), confirming the two metrics capture unrelated constructs.

\subsection{Circularity: Why Profile Extraction ``Wins'' on the Judge}
\label{sec:circularity}

The discrepancy has a concrete explanation.
\method{Profile Extraction} achieves TMR$=$0.542, far above other methods, but LUAR$=$0.502---indistinguishable.
Both the judge's trait extraction (Stage~1) and the method's profile extraction ask an LLM to read author samples and extract salient style features.
The profile is then used to generate text optimized for exactly the kind of features the judge checks.

The real author's own text confirms this: TMR$=$0.427, \emph{lower} than \method{Profile Extraction}'s 0.542.
If TMR measured genuine authorship fidelity, the real author would set the ceiling.
Instead, the method that explicitly optimizes for LLM-extractable traits exceeds the real author, exposing that TMR measures \textbf{instruction-following fidelity}, not \textbf{authorship fidelity}.

Additionally, we find trait extraction is unstable: repeated extraction for the same author yields mean Jaccard similarity of 0.22 across trait sets.
The yardstick itself changes between measurements.

\subsection{Cross-Model Robustness}

To test whether the authorship gap is generator-specific, we replicate with GLM-4 32B (10 authors, 150 generations).
All GLM-4 methods score below the human floor: LUAR ranges 0.417--0.576 vs.\ floor 0.626.
GLM-4 shows wider method spread (0.16 vs.\ Qwen's 0.024) but the gap persists across both model families.

Cross-model LUAR analysis reveals three distinct regimes: within-model similarity is high (Qwen$\leftrightarrow$Qwen 0.918, GLM$\leftrightarrow$GLM 0.839), cross-model is lower (Qwen$\leftrightarrow$GLM 0.753), and gen$\rightarrow$real is lowest (0.45--0.49).
Each LLM carries its own authorship fingerprint that inference-time personalization does not erase---a finding consistent with theoretical predictions from AI-generated text detection~\citep{mitchell2023detectgpt}.

\subsection{Prompt Contamination as Confound}

A methodological finding relevant to benchmark design: na\"ive prompt construction (extracting the first sentence of the target post) inflates the unpersonalized baseline from SA$=$22\% to 50\%---a \textbf{28 percentage point} confound.
The first sentence carries the author's vocabulary, punctuation, and tonal markers, making even unpersonalized output appear author-matched.
We adopt LLM-extracted content summaries following \citet{kumar2024longlamp}, which specify \emph{what} the post discusses without leaking \emph{how} the author writes.
Prompt construction is a hidden confound that can mask or invert method rankings.

\section{Discussion}
\label{sec:discussion}

\paragraph{Design principles from the theory--benchmark cycle.}
Our findings yield two design principles for stylistic personalization evaluation:

\emph{(i) Ground metrics in established theory.}
Authorship verification provides what ad hoc metrics cannot: calibrated baselines with absolute meaning.
LUAR's reference points (ceiling: 0.756, floor: 0.626) transform a score of 0.50 from uninterpretable to precise---all methods remain 0.12--0.14 below even the cross-author floor, let alone the same-author ceiling.
Repurposing task-accuracy or preference-alignment metrics for stylistic fidelity fails because they measure different constructs entirely.

\emph{(ii) Treat multi-metric disagreement as diagnostic.}
Rather than averaging metrics or picking the most convenient one, disagreement between independently motivated metrics should be treated as evidence of construct validity failure~\citep{cronbach1955construct}---a signal that some metrics are measuring artifacts rather than the target construct.

\paragraph{Toward the theory--benchmark cycle.}
The CTB vision of a virtuous cycle between theory and benchmarks applies directly: authorship verification theory predicts that LLM-generated text carries a model-specific fingerprint detectable by trained classifiers~\citep{mitchell2023detectgpt, kirchenbauer2023watermark}.
Our LUAR measurements confirm this prediction quantitatively: the fingerprint is strong enough that all personalized outputs cluster below the human floor.
This provides a \emph{falsifiable, calibrated} measure of personalization capability---exactly the kind of formal guarantee the field needs to move beyond ad hoc evaluation.
Concretely, our framework instantiates two legs of the cycle: theory (authorship verification) generates a testable prediction (inference-time prompting cannot shift the LLM's fingerprint toward a target author), and the benchmark confirms it with calibrated measurements, identifying the precise gap future methods must close.

\paragraph{The generated-text regime.}
Our analysis reveals a structural phenomenon: LLM-generated text occupies a distinct region of LUAR embedding space.
Gen$\leftrightarrow$gen similarity averages 0.932 (same target) vs.\ 0.858 (different targets), both far above gen$\leftrightarrow$real (0.522).
Personalization modulates output \emph{within} this regime (gen$\leftrightarrow$gen AUC$=$0.918) but does not escape it.
This suggests a formal characterization may be possible: the set of achievable authorship embeddings under inference-time methods may be bounded away from the human manifold, providing a theoretical target for future work on closing the gap.

\paragraph{Scope.}
This study targets inference-time stylistic personalization (few-shot, profile extraction, contrastive examples) on blog writing from one corpus in English using two model families (Qwen~3, GLM-4) at 32B scale.
Training-time approaches such as per-user LoRA adapters or preference optimization may shift the LLM's authorship fingerprint more substantially; this work provides the calibrated yardstick on which they should be judged.
Analogous inference-time personalization failures have been observed in behavioral domains~\citep{sawant2026highstakes}, suggesting the authorship gap may extend beyond stylistic fidelity to other personalization constructs.

\paragraph{Does LUAR penalize generated text per se?}
A natural concern is that LUAR's low gen$\rightarrow$real similarity reflects generic ``LLM-ness'' detection rather than author-style mismatch.
Three observations weigh against this reading.
First, LUAR achieves AUC$=$0.918 at distinguishing target authors \emph{within} generated text: if it merely detected machine origin, all generations would cluster identically regardless of target.
Second, cross-model LUAR exhibits graded structure (Qwen$\leftrightarrow$Qwen 0.918, Qwen$\leftrightarrow$GLM 0.753, gen$\rightarrow$real 0.45--0.49), consistent with author-style \emph{mixing with} model-style rather than a binary human/machine cut.
Third, GLM-4 produces a method spread of 0.16 in LUAR (vs.\ Qwen's 0.024), showing LUAR registers method-level differentiation \emph{inside} the generation regime.
LUAR thus measures a continuous authorship signal that is partially recoverable under inference-time prompting and partially captured by the model's own fingerprint.

\paragraph{Future work.}
Three complementary directions follow.
(i) Human evaluation of perceived stylistic fidelity, to triangulate which metric best predicts reader-perceived authorship match.
(ii) Training-time methods (per-user LoRA, preference optimization) to test whether the gap closes when the generator's fingerprint can be reshaped, rather than only its conditional distribution.
(iii) Cross-genre validation (microblog, email, code commentary, scientific writing) to test whether the gap and its calibrated baselines transfer across communicative settings.
All evaluation is in English; authorship patterns may differ across languages.


\bibliography{references}

@inproceedings{salemi2024lamp,
  title={{LaMP}: When Large Language Models Meet Personalization},
  author={Salemi, Alireza and Mysore, Sheshera and Bendersky, Michael and Zamani, Hamed},
  booktitle={ACL},
  year={2024}
}

@inproceedings{personalllm2025,
  title={{PersonalLLM}: Tailoring {LLM}s to Individual Preferences},
  author={Zollo, Thomas P and Siah, Kwan Ho and Ye, Tian and Li, Hurui and Namkoong, Hongseok},
  booktitle={ICLR},
  year={2025}
}

@inproceedings{zhao2025personalens,
  title={{PersonaLens}: A Benchmark for Personalization Evaluation in Conversational {AI} Assistants},
  author={Zhao, Zheng and Vania, Clara and Kayal, Subhradeep and Khan, Naila and Cohen, Shay B and Yilmaz, Emine},
  booktitle={Findings of ACL},
  year={2025}
}

@article{cronbach1955construct,
  title={Construct validity in psychological tests},
  author={Cronbach, Lee J and Meehl, Paul E},
  journal={Psychological Bulletin},
  volume={52},
  number={4},
  pages={281--302},
  year={1955}
}

@inproceedings{riverasoto2021luar,
  title={Learning Universal Authorship Representations},
  author={Rivera-Soto, Rafael and Miano, Olivia and Ordonez, Juanita and Chen, Barry Y and Khan, Aleem and Bishop, Marcus and Andrews, Nicholas},
  booktitle={EMNLP},
  year={2021}
}

@inproceedings{schler2006effects,
  title={Effects of Age and Gender on Blogging},
  author={Schler, Jonathan and Koppel, Moshe and Argamon, Shlomo and Pennebaker, James W},
  booktitle={AAAI Spring Symposium on Computational Approaches to Analyzing Weblogs},
  year={2006}
}

@article{qwen2025qwen3,
  title={Qwen3 Technical Report},
  author={{Qwen Team}},
  journal={arXiv preprint},
  year={2025}
}

@article{glm2024chatglm,
  title={{ChatGLM}: A Family of Large Language Models from {GLM-130B} to {GLM-4} All Tools},
  author={{GLM Team}},
  journal={arXiv preprint arXiv:2406.12793},
  year={2024}
}

@article{argamon2003style,
  title={Gender, Genre, and Writing Style in Formal Written Texts},
  author={Argamon, Shlomo and Koppel, Moshe and Fine, Jonathan and Shimoni, Anat Rachel},
  journal={Text \& Talk},
  volume={23},
  pages={321--346},
  year={2003}
}

@article{stamatatos2009survey,
  title={A Survey of Modern Authorship Attribution Methods},
  author={Stamatatos, Efstathios},
  journal={Journal of the American Society for Information Science and Technology},
  volume={60},
  number={3},
  pages={538--556},
  year={2009}
}

@inproceedings{mitchell2023detectgpt,
  title={{DetectGPT}: Zero-Shot Machine-Generated Text Detection Using Probability Curvature},
  author={Mitchell, Eric and Lee, Yoonho and Khazatsky, Alexander and Manning, Christopher D and Finn, Chelsea},
  booktitle={ICML},
  year={2023}
}

@inproceedings{kirchenbauer2023watermark,
  title={A Watermark for Large Language Models},
  author={Kirchenbauer, John and Geiping, Jonas and Wen, Yuxin and Katz, Jonathan and Miers, Ian and Goldstein, Tom},
  booktitle={ICML},
  year={2023}
}

@article{kumar2024longlamp,
  title={{LongLaMP}: A Benchmark for Personalized Long-Form Text Generation},
  author={Kumar, Saket and Sathe, Chinmay and Tiwari, Ashutosh and Zamani, Hamed},
  journal={arXiv preprint arXiv:2407.11016},
  year={2024}
}

@inproceedings{betterbench2024,
  title={{BetterBench}: Assessing {AI} Benchmarks, Uncovering Issues, and Establishing Best Practices},
  author={Reuel, Anka and Hardy, Amelia and Lamparth, Max and Hardy, Mitchell and Smith, Bernease and Kochenderfer, Mykel J},
  booktitle={NeurIPS Datasets and Benchmarks},
  year={2024}
}

@inproceedings{raji2021benchmark,
  title={{AI} and the Everything in the Whole Wide World Benchmark},
  author={Raji, Inioluwa Deborah and Bender, Emily M and Paullada, Amandalynne and Denton, Emily and Hanna, Alex},
  booktitle={NeurIPS},
  year={2021}
}

@inproceedings{wang2025catchme,
  title={Catch Me If You Can? {Not Yet}: {LLM}s Still Struggle to Imitate the Implicit Writing Styles of Everyday Authors},
  author={Wang, Sheng and others},
  booktitle={EMNLP Findings},
  year={2025}
}

@article{sawant2026highstakes,
  title={High-Stakes Personalization: Rethinking {LLM} Customization for Individual Investor Decision-Making},
  author={Sawant, Yash Ganpat},
  journal={arXiv preprint arXiv:2604.04300},
  year={2026}
}
\bibliographystyle{icml2026}

\end{document}